\pdfoutput=1

\documentclass[11pt]{article}

\usepackage[]{ACL2023}

\usepackage{times}
\usepackage{latexsym}

\usepackage[T1]{fontenc}

\usepackage[utf8]{inputenc}

\usepackage{microtype}

\usepackage{inconsolata}
\usepackage{graphicx}
\usepackage{amssymb}
\usepackage{makecell}
\usepackage{comment}
\usepackage{amsmath}
\usepackage{hyperref}
\title{Named Entity Recognition via Machine Reading Comprehension: A Multi-Task Learning Approach}

\author{Yibo Wang \\
  Department of Computer Science \\
  University of Illinois Chicago  \\
  \texttt{ywang633@uic.edu} \\\And
  Wenting Zhao \\
  Department of Computer Science \\
  University of Illinois Chicago \\
  \texttt{wzhao41@uic.edu} \\\AND
  Yao Wan \\
  School of Computer Science and Technoloy \\
  Huazhong University of Science and Technology\\
  \texttt{wanyao@hust.edu.cn} \\\And
  Zhongfen Deng \\
  Department of Computer Science \\
  University of Illinois Chicago \\
  \texttt{zdeng21@uic.edu} \\\AND
  Philip S. Yu \\
  Department of Computer Science \\
  University of Illinois Chicago \\
  \texttt{psyu@uic.edu} \\}

\begin{document}
\maketitle
\begin{abstract}
Named Entity Recognition (NER) aims to extract and classify entity mentions in the text into pre-defined types (e.g., organization or person name).
Recently, many works have been proposed to shape the NER as a machine reading comprehension problem (also termed MRC-based NER), in which entity recognition is achieved by answering the formulated questions related to pre-defined entity types through MRC, based on the contexts.
However, these works ignore the label dependencies among entity types, which are critical for precisely recognizing named entities.
In this paper, we propose to incorporate the label dependencies among entity types into a multi-task learning framework for better MRC-based NER.
We decompose MRC-based NER into multiple tasks and use a self-attention module to capture label dependencies.
Comprehensive experiments on both nested NER and flat NER datasets are conducted to validate the effectiveness of the proposed Multi-NER.
Experimental results show that Multi-NER can achieve better performance on all datasets.
\end{abstract}

\section{Introduction}
\begin{figure}[t]
    \centering
    \includegraphics[width=0.48\textwidth]{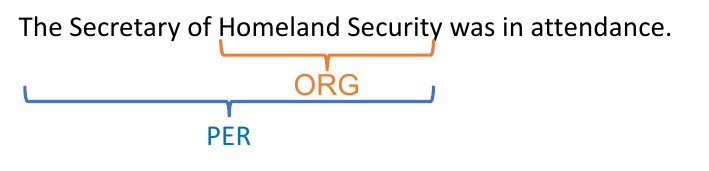}
    \caption{An example of nested NER.}
    \label{fig:example}
\end{figure}

Named Entity Recognition (NER), which aims to locate and classify entity mentions in text into pre-defined types, is a fundamental task in information extraction~\cite{chinchor1997muc, nadeau2007survey}. 
Typically, NER is formulated as a sequence labeling task, where each token is classified as one of the pre-defined types.
However, the sequence labeling models can only assign one label to a token, resulting in the incapability of handling overlapping entities in nested NER
~\cite{finkel2009nested}.
Figure~\ref{fig:example} shows an example of nested NER: \textit{Homeland Security} can be recognized as 
\textsc{Organization}
, as well as \textsc{Person}. 

To mitigate this issue, many works resort to formulating NER as a Machine Reading Comprehension (MRC) question answering task (termed MRC-based NER)~\cite{wang2020learning, li2020unified,wang2022continuous}. For example, to recognize the \textsc{Organization}, a natural-language question \textit{``Which \textsc{Organization} is mentioned in the text?''} is formulated. 
Then the goal of NER is transformed to answer the formulated questions through machine reading comprehension, given the contexts.
MRC-based NER provides a unified solution for both flat and nested NER tasks since 
each entity type has its corresponding entity span positions as the answer, and these output answers are independent of each other.

Despite much progress having been made in MRC-based NER, existing approaches tend to ignore the label dependencies among entity types, which are critical for precise NER.
Label dependencies indicate that different entities in the text have some relations with each other. 
For example, in a sentence \textit{``Ousted WeWork founder Adam Neumann lists his Manhattan penthouse for \$37.5 million''}, once \textit{Adam Neumann} is recognized as
\textsc{Person},
it is expected to help with the recognition of \textit{WeWork} as \textsc{Organization} because the \textit{founder} preceding a person's name implies an organization.

\begin{figure}[t]
	\centering
	\includegraphics[width=0.5\textwidth]{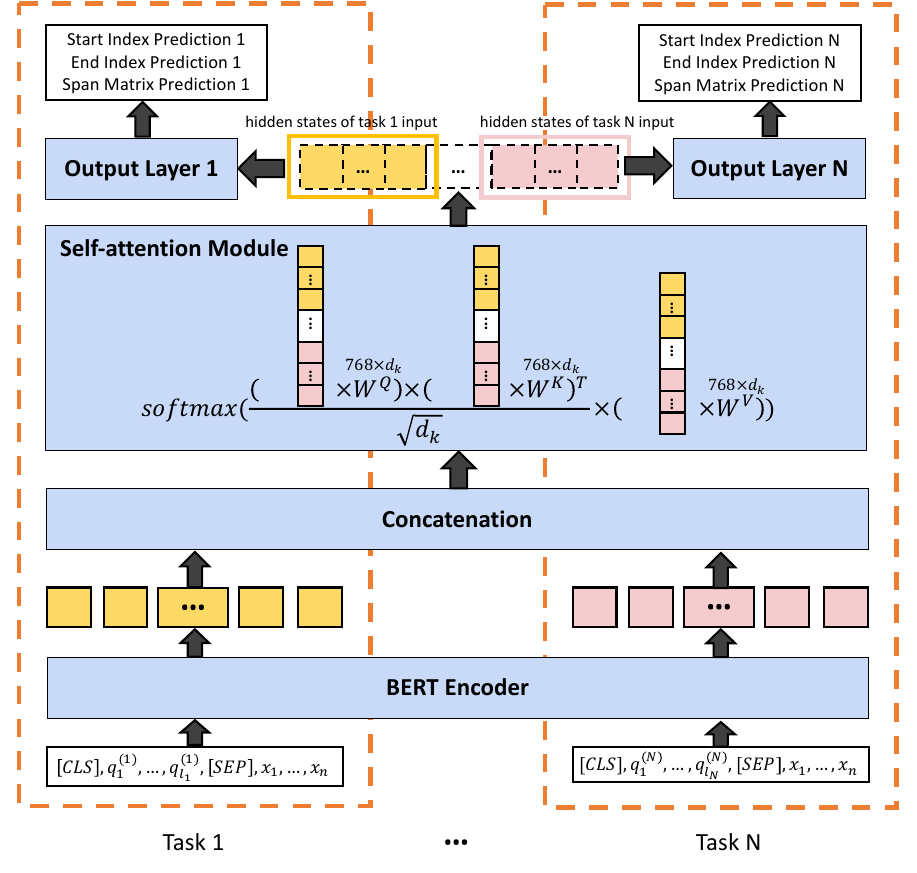}
	\caption{The architecture of our proposed Multi-NER. 
 The input of each task is the concatenation of a question associated with task/entity type and context. The outputs are the corresponding start index prediction, end index prediction, and span matrix prediction.
 }
	\label{fig:1}
\end{figure}

To leverage the label dependencies among entity types, we propose a novel multi-task learning framework (termed Multi-NER) for MRC-based NER. 
In Multi-NER, 
MRC-based NER is decomposed into multiple tasks, each task focusing on one entity type. For each task,
the corresponding input is
the concatenation of an entity-class-related question and the context,
and the output is expected to be the corresponding entity spans (i.e., start and end positions). 
The input is first encoded
via a pre-trained BERT~\cite{devlin-etal-2019-bert}.
The concatenation of embeddings of all tasks are fed into a self-attention module, which can preserve the 
label dependencies between different entity types.
Finally, task-specific output layers are applied to different tasks.

To validate the effectiveness of our proposed Multi-NER, we conduct experiments on the datasets of both flat NER and nested NER.
Experimental results show that Multi-NER can benefit the MRC-based NER in both flat NER and nested NER, with the important label dependencies among entities preserved. 
Additionally, we also visualize the self-attention maps to examine whether the label dependencies have been successfully captured. 



Overall, the contributions of this paper are two-fold: 1) We are the first to propose a multi-task learning framework for MRC-based NER tasks to capture label dependencies between entity types; 
2) The introduced self-attention maps are visualized to verify that the self-attention modules can capture label dependencies.

All the source code and datasets are available at \href{https://github.com/YiboWANG214/MultiNER}{https://github.com/YiboWANG214/MultiNER}

\section{Methodology}

\subsection{Problem Formulation}

Given a sequence $X=\{x_1, x_2, \ldots, x_n\}$, where $n$ denotes length of $X$, NER aims to find every entity mention in $X$ and assign an entity type $y\in\mathcal{Y}$ to it. BERT-MRC \cite{li2020unified} transforms tagging-style NER to MRC format with a triplet of (\textsc{Question}, \textsc{Answer}, \textsc{Context}). 
The natural language question $q^{(y)} = \{q^{(y)}_1,\ldots,q^{(y)}_{l_y}\}$, where $l_y$ denotes length of $q^{(y)}$, is related to the entity type $y$ and considered as \textsc{Question};
The positions $P^y_{start,end}$ of entity mentions of $y$ is considered as \textsc{Answer};
The input sequence $X$ is considered as \textsc{Context}. Given $X$ and $q^{(y)}$, the goal of BERT-MRC is to predict $P^y_{start,end}$.

Our Multi-NER applies the same MRC format but further decomposes BERT-MRC into multiple tasks, where each task $i\in\{1, 2, \ldots, |\mathcal{Y}|\}$ only focuses on one entity type. Thus, the task set $\{1, 2, \ldots, |\mathcal{Y}|\}$ and the entity type set $\mathcal{Y}$ are bijection.
Instead of processing one \textsc{Question} at a time, Multi-NER processes all \textsc{Question}s in a multi-task framework.
Therefore, in Multi-NER settings, given a \textsc{Context} $X$ and multiple questions $\{q^{(y)}\}_{y\in\mathcal{Y}}$, the goal is to predict $\{P^y_{start, end}\}_{y\in\mathcal{Y}}$.

\subsection{Multi-NER}
Figure~\ref{fig:1} gives an overview of our proposed Multi-NER, which
consists of $N$ tasks, 
where each task denotes the recognition of one specific entity type.
To share information between tasks, one shared encoder is used across tasks, and a self-attention module is employed to capture label dependencies.



\paragraph{Single Task Learning}
For every single task $i\in\{1, 2, \ldots, |\mathcal{Y}|\}$, 
the input sequence is the concatenation of the natural language question $q^{(i)}$ and the context $X$, as follows:
\begin{equation}
    I^{(i)} = \texttt{[CLS]}, q^{(i)}, \texttt{[SEP]}, X\,.
\end{equation}
The output of task $i$ has three components: start index prediction, end index prediction, and span matrix prediction. The start index prediction is the probability of each token being a start position; the end index prediction is the probability of each token being an end position; the span matrix prediction is the probability of each start-end pair being an entity mention position.

\paragraph{Task Interactions}
Task interactions are two-fold. First, one shared large language model like BERT~\cite{devlin-etal-2019-bert} is used as the encoder for all tasks to make the embedding space consistent. Thus, the embedding of task $i$ is $E^{(i)} = Encoder(I^{(i)})$.
Second, a self-attention module~\cite{vaswani2017attention} is used across all tasks, accepting the concatenation of the embeddings of every task as input. The self-attention module incorporates information from every task and outputs the concatenation of hidden states
\begin{equation}
    E = ATT(\left[E^{(1)}, \cdots, E^{(|\mathcal{Y}|)}\right])\,.
\end{equation}
The self-attention module enables $E$ with the property of capturing the label dependencies between entity types, making $E$ a better representation.

\paragraph{Model Learning}
At training time, all tasks are trained jointly. The loss function for multi-task learning is defined as follows:
\begin{equation}
    \mathcal{L} = \sum_{i=1}^{|\mathcal{Y}|} (\alpha\mathcal{L}^i_{start} + \beta\mathcal{L}^i_{end}  + \gamma\mathcal{L}^i_{span} ),
\end{equation}
where $\alpha$, $\beta$, and $\gamma$ are tunable weights for start index prediction, end index prediction, and span matrix prediction. $\mathcal{L}^i_{start}$, $\mathcal{L}^i_{end}$ and $\mathcal{L}^i_{span}$ are cross entropy loss of task $i$.

\section{Experiments}
To evaluate the performance of the proposed Multi-NER, we compare it with a 
state-of-the-art 
baseline BERT-MRC \cite{li2020unified}, on 
datasets of flat NER and nested NER.
We also perform a case study with attention maps visualized to further analyze the ability of Multi-NER to capture label dependencies.

\subsection{Datasets}
We adopt three nested NER datasets (i.e., English ACE-2004 \cite{mitchell2005ace}, English ACE-2005 \cite{walker2006ace} and GENIA \cite{ohta2002genia}) and one flat NER dataset (i.e., English CoNLL-2003 \cite{sang2003introduction}) to evaluate the performance of Multi-NER. 
ACE-2004 and ACE-2005 are two textual datasets from broadcast, newswire, telephone conversations and weblogs. 
GENIA is a collection of biomedical literature, containing Medline
abstracts. 
CoNLL-2003 is extracted from Reuters news stories between August 1996 and August 1997. 
We conducted experiments on both nested NER datasets and flat NER dataset since our model is based on BERT-MRC, which can be applied to both flat and nested NER.

\subsection{Experimental Settings}
For a fair comparison,
we select $BERT_{base}$ as the backbone encoder for all models. 
We adopt a one-layer linear transformation to predict the start index and end index and adopt a two-layer MLP with activation function GELU to predict the span matrix.
We set the hidden size as 1,536 and the dropout rate as 0.1.
More details of the hyperparameters setting are referred to the Appendix~\ref{sec:hyperparameters}.
We follow the same process of question generation in~\cite{li2020unified} to use annotation guideline notes, which are the guidelines provided to the annotators when building datasets, as references to construct questions.

\subsection{Results and Analysis}
Table~\ref{tab:results} shows the experimental results on both nested NER datasets and flat NER dataset. From this table, we can observe that our Multi-NER achieves 85.34\% on ACE 2004, 84.25\% on ACE 2005, 81.13\% on GENIA, and 92.33\% on CoNLL-2003, achieving +1.3\%, +0.4\%, +1.24\% and 1.25\% improvement, respectively, when comparing with BERT-MRC. 
The performance improvements of Multi-NER on all the datasets indicate that formulating MRC-based NER into a multi-task learning framework to obtain label dependencies between different entity types can indeed bring model performance improvement.

\begin{table}[t]
    \centering
    \caption{Experimental results on nested NER datasets and flat NER dataset.}
     \setlength{\tabcolsep}{4pt} 
    \begin{tabular}{l|ccc}
        \hline
        \textbf{Model} & \textbf{Precision} & \textbf{Recall} & \textbf{F1} \\
        \hline
        \multicolumn{4}{c}{ACE 2004} \\
        \hline
        BERT-MRC & 84.63 & 83.46 & 84.04 \\
        Multi-NER & 86.01 & 84.68 & 85.34 (+1.3)\\
        \hline
        \multicolumn{4}{c}{ACE 2005} \\
        \hline
        BERT-MRC & 83.22 & 84.48 & 83.85 \\
        Multi-NER & 84.47 & 84.03 & 84.25 (+0.4) \\
        \hline
        \multicolumn{4}{c}{GENIA} \\
        \hline
        BERT-MRC & 79.47 & 80.32 & 79.89 \\
        Multi-NER & 82.63 & 79.68 & 81.13 (+1.24)\\
        \hline
        \multicolumn{4}{c}{CoNLL-2003} \\
        \hline
        BERT-MRC & 90.61 & 91.55 & 91.08 \\
        Multi-NER & 91.96 & 92.70 & 92.33 (+1.25)\\
        \hline
    \end{tabular}
    \label{tab:results}
\end{table}

Furthermore, the idea behind our proposed multi-task framework is to use different output layers to disambiguate between entity types and use a self-attention module to obtain label dependencies between entity types. 
To evaluate the contribution of the different output layers and the self-attention module, we also conduct ablation studies on all datasets. 
The experimental results in Table~\ref{tab:ablation} show that both 
different output layers and the self-attention module
contribute to Multi-NER.

\begin{table}[]
    \centering
    \caption{Ablation studies evaluate the contribution of components of Multi-NER.}
     \setlength{\tabcolsep}{3pt} 
    \resizebox{\linewidth}{!}{
    \begin{tabular}{l|cccc}
    \hline
        \textbf{Model} & \textbf{ACE04} & \textbf{ACE05} & \textbf{GENIA} & \textbf{CoNLL03} \\
        \hline
        Multi-NER & 85.34 & 84.25 & 81.13 & 92.33 \\
        (w/o att) & 84.84 & 84.01 & 80.87 & 91.32 \\
        (w/o diff) & 84.72 & 84.05 & 80.58 & 91.55 \\
        \hline
    \end{tabular}
    }
    \label{tab:ablation}
\end{table}

\subsection{Case Study}
\begin{figure}
    \centering
    \includegraphics[width=0.48\textwidth]{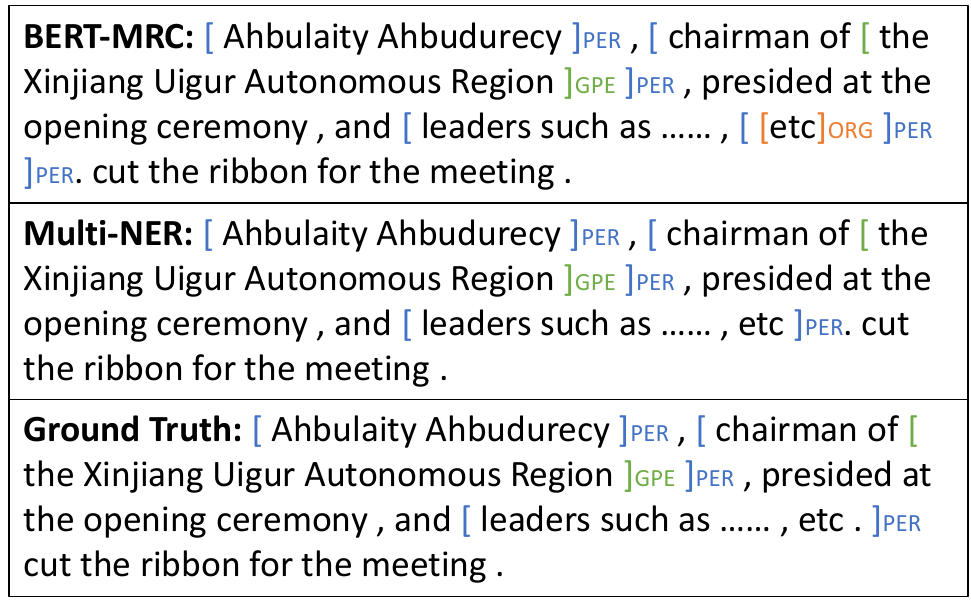}
    \vspace{-2mm}
    \caption{The ground truth, predictions of BERT-MRC and Multi-NER for a randomly selected example. The middle of the sentence is omitted for better presentation. The complete example is referred to Appendix~\ref{sec:examples}.}
    \label{fig:case}
\end{figure}

To further study the effect of Multi-NER, we examine some randomly selected examples.
As an example, for \textit{``Ahbulaity Ahbudurecy , chairman of the Xinjiang Uigur Autonomous Region , presided at the opening ceremony , and leaders such as Tiemuer Dawamaiti , vice - chairman of the Standing Committee of the National People ' s Congress , Shimayi Aimaiti , member of the State Affairs Committee , etc . cut the ribbon for the meeting .''} from ACE 2004, we show the ground-truth entities and predicted results of BERT-MRC and Multi-NER in Figure~\ref{fig:case}.
In BERT-MRC, \textit{etc} is categorized as both \textsc{ORG} and \textsc{PER}, while in Multi-NER and ground truth \textit{etc} is categorized as \textsc{PER}. 
Ambiguous tokens like \textit{etc} are hard to categorize even with contextual information. However, when applying Multi-NER, different entity types and label dependencies are also considered, which is beneficial to ambiguous tokens.

\begin{figure}[!t]
    \centering
	\includegraphics[width=0.4\textwidth]{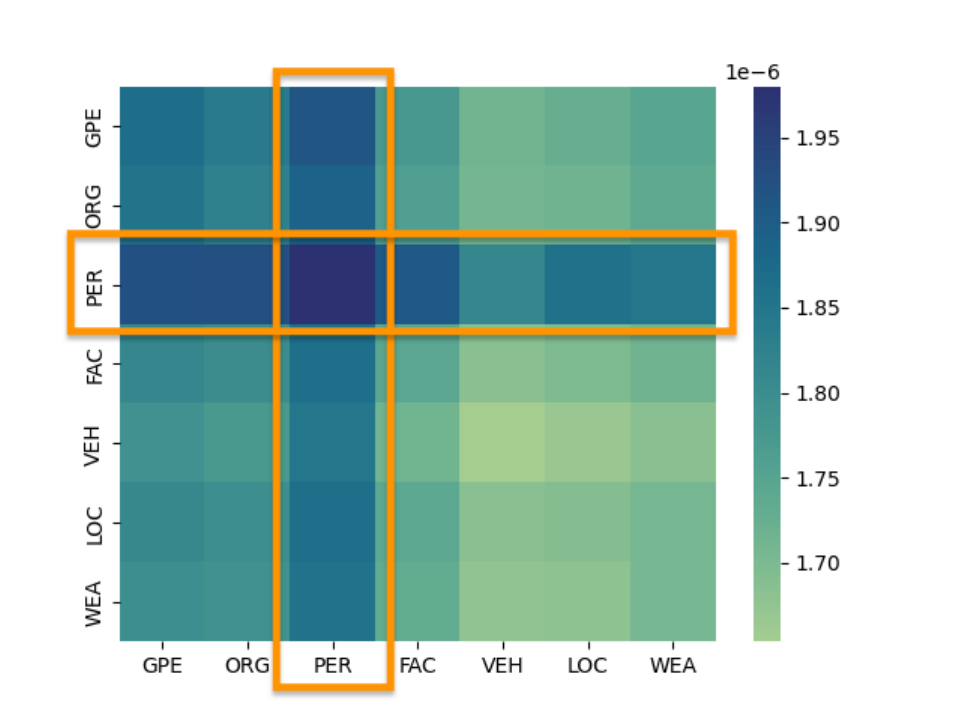}
	\caption{Attention map of the mean scores according to entity types of a randomly selected example from ACE 2004. }
	\label{fig:attention}
\end{figure}

We also show the attention map of the mean scores according to entity types of this example in Figure~\ref{fig:attention}.
We can see that \textsc{PER} has a relatively large impact on other entity types, helping the model improve performance on other entity types with \textsc{PER} information.
We attribute it to label dependencies obtained by information sharing between entity types using the self-attention module.

\section{Related Work}
As language models have advanced \cite{devlin-etal-2019-bert, t5, dong2023closed, zhao2021attend}, numerous efforts have emerged to enhance the performance of MRC-based NER.
\citet{zhang2022multi} incorporated different domain knowledge into MRC-based NER task to improve model generalization ability. 
\citet{liu2022fusing} proposed to use graph attention networks to capture label dependencies between entity types when applying MRC-based NER to electronic medical records. However, they only use entity type embeddings to build graph attention networks, ignoring the rich information in context.
MRC-based NER is applied to different domains. 
\citet{du2022mrc} designed an MRC-based method for medical NER through both sequence labeling and span boundary detection.
\citet{zhang2022finbert} applied MRC-based NER for financial named entity recognition from literature.
\citet{wang2020learning} proposed MRC-based NER with the help of a distilled masked language model in e-commerce. 
\citet{jia2022query} applied MRC-based methods for multimodal named entity recognition.

The span-based methods \cite{eberts2020span} that formulate nested NER as a span classification task are also mentionable.  
\citet{wan2022nested} improved span representation using retrieval-based span-level graphs based on n-gram similarity. 
\citet{yuan2022fusing} integrated heterogeneous factors like inside tokens, boundaries, labels, and related spans to improve the performance of span representation and classification.
\citet{shen2021locate} improved span-based NER using a two-stage entity identifier to filter out low-quality spans to reduce computational costs. 

\section{Conclusion}
In this paper, we propose to incorporate the label dependencies among entity types into a novel multi-task learning framework (termed Multi-NER) for MRC-based NER. 
A self-attention mechanism is introduced to obtain label dependencies between entity types. 
Experimental results validate that Multi-NER outperforms BERT-MRC on both nested NER and flat NER. 
Case study and attention map visualization show that our introduced self-attention module is able to capture label dependencies among entities, contributing to a performance improvement.

\section*{Limitations}
One limitation of our proposed Multi-NER lies in that the number of tasks depends on the number of entity types because each entity type is considered as a task. 
Depending on the model structure we use in Multi-NER, the number of parameters will increase by 4M for each additional entity type with $Max\_Length=128$. 
One possible solution to solve this problem is using parameter-efficient fine-tuning methods like Hypernetworks~\cite{DBLP:journals/corr/HaDL16} to effectively generate task-specific output layers.
We leave this problem to our future work.

\bibliography{anthology,custom}
\bibliographystyle{acl_natbib}

\appendix

\section{Appendix}
\label{sec:appendix}

\subsection{Hyperparameters}
\label{sec:hyperparameters}
The hyperparameter details are shown in Table \ref{tab:hyperparameters}.

\begin{table*}[h]
    \centering
    \caption{Hyperparameters for all models.}
    \begin{tabular}{cccccc}
        \hline
        \multicolumn{6}{c}{ACE 2004} \\
        \hline
        Model & Batch Size & Max Length & Learning Rate & Epoch & \# Parameters\\
        BERT-MRC & 2 & 128 & 3e-5 & 14 & 112M \\
        Multi-NER & 2 & 128 & 2e-5 & 19 & 133M \\
        \hline
        \multicolumn{6}{c}{ACE 2005} \\
        \hline
        Model & Batch Size & Max Length & Learning Rate & Epoch & \# Parameters \\
        BERT-MRC & 2 & 128 & 2e-5 & 11 & 112M \\
        Multi-NER & 2 & 128 & 2e-5 & 16 & 133M \\
        \hline
        \multicolumn{6}{c}{GENIA} \\
        \hline
        Model & Batch Size & Max Length & Learning Rate & Epoch & \# Parameters \\
        BERT-MRC & 2 & 180 & 2e-5 & 9 & 112M \\
        Multi-NER & 2 & 128 & 2e-5 & 15 & 125M \\
        \hline
        \multicolumn{6}{c}{CoNLL-2003} \\
        \hline
        Model & Batch Size & Max Length & Learning Rate & Epoch & \# Parameters \\
        BERT-MRC & 2 & 200 & 3e-5 & 8 & 112M \\
        Multi-NER & 1 & 200 & 2e-5 & 18 & 123M \\
        \hline
    \end{tabular}
    \label{tab:hyperparameters}
\end{table*}

\subsection{Examples}
\label{sec:examples}
\begin{figure*}
    \centering
    \includegraphics[width=1\textwidth]{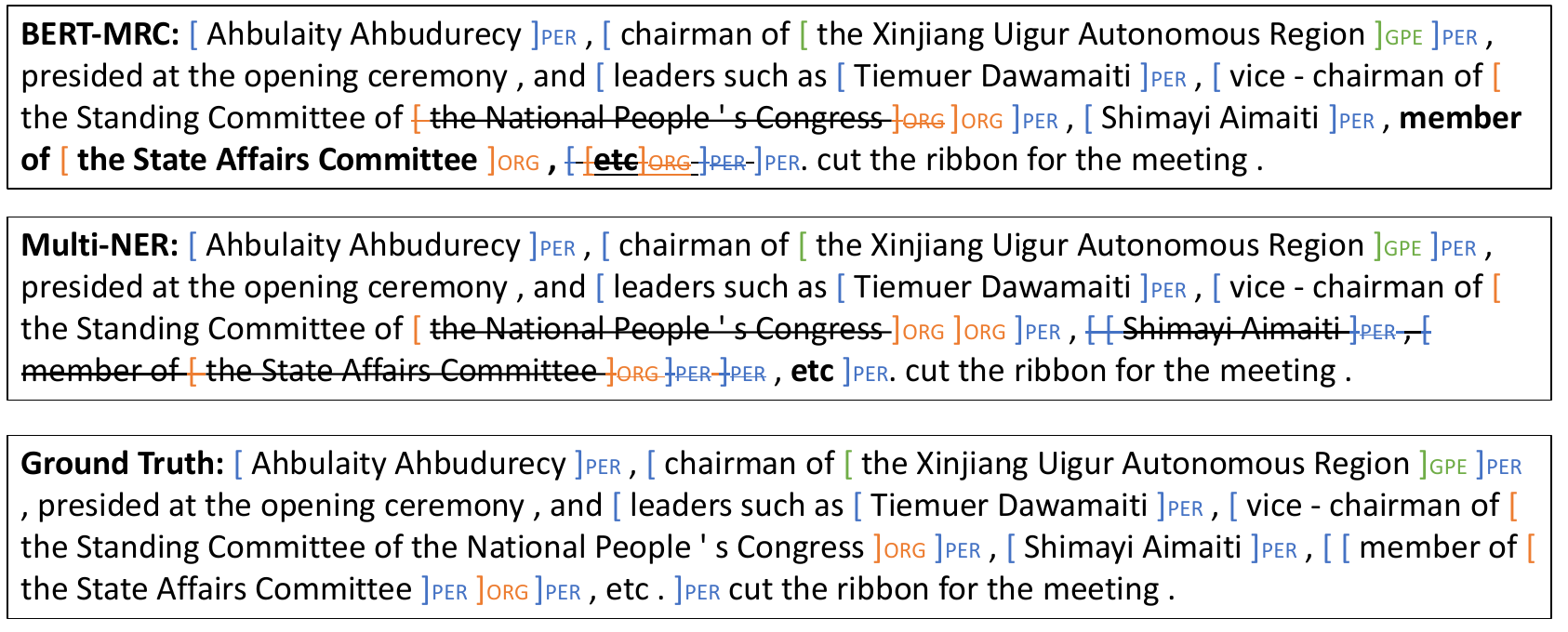}
    \caption{The ground truth, results of BERT-MRC and Multi-NER of a randomly selected example. Delete lines and underlines represent false positive and bolded tokens represent false negative. Punctuation errors are not indicated in the figure.}
    \label{fig:case_full}
\end{figure*}

The ground truth, predicted results of Multi-NER and BERT-MRC of a randomly selected example are shown in Table~\ref{tab:example} and Figure~\ref{fig:case_full}.
From the table, we can observe that the true positive, false positive, and false negative positions of BERT-MRC and Multi-NER are 8, 4, 3, and 9, 3, 2. Besides, most errors of Multi-NER are due to the ambiguity of the entity mention boundaries like punctuation.

\begin{table*}[h]
    \centering
    \caption{Results of an example \textit{``Ahbulaity Ahbudurecy , chairman of the Xinjiang Uigur Autonomous Region , presided at the opening ceremony , and leaders such as Tiemuer Dawamaiti , vice - chairman of the Standing Committee of the National People ' s Congress , Shimayi Aimaiti , member of the State Affairs Committee , etc . cut the ribbon for the meeting .''} from ACE 2004.}
    \begin{tabular}{cccc}
        \hline
         entity type & BERT-MRC & Multi-NER & Ground Truth\\
         \hline
         GPE & (5,9) & (5,9) & (5,9) \\
         \hline
         ORG & \makecell[c]{(28,37) (32,37)\\ (44,47) (49,49)} & \makecell[c]{(28,37) (32,37)\\ (44,47)} & \makecell[c]{(28,37) (44,47)}\\
         \hline
         PER & \makecell[c]{(0,1) (3,9) (18,49)\\ (21,22)  (24,37)\\ (39,40) (49,49)} & \makecell[c]{(0,1) (3,9) (18,49)\\ (21,22)  (24,37)\\ (39,40) (39,47) (42,47) }  & \makecell[c]{(0,1) (3,9) (18,50)\\ (21,22)  (24,37)\\ (39,40) (42,47) (49,50)} \\
         \hline
         FAC & - & - & - \\
         \hline
         VEH & - & - & - \\
         \hline
         LOC & - & - & - \\
         \hline
         WEA & - & - & - \\
         \hline
    \end{tabular}
    \label{tab:example}
\end{table*}


\end{document}